# Reconstruction and Registration of Large-Scale Medical Scene Using Point Clouds Data from Different Modalities

Ke Wang   Han Song   Jiahui Zhang   Xinran Zhang   Hongen Liao*

Department of Biomedical Engineering, School of Medicine, Tsinghua University
Beijing, China

k-wan14@mails.tsinghua.edu.cn   songh17@mails.tsinghua.edu.cn

jiahui-z15@mails.tsinghua.edu.cn   zhangxinran@tsinghua.edu.cn   liao@tsinghua.edu.cn

## 1. Introduction

Sensing the medical scenario can ensure the safety during the surgery operations. So, in this regard, a monitor platform which can obtain the accurate location information of the surgery room is desperately needed. Compared to 2D camera image, 3D data contains more information of distance and direction. Therefore, 3D sensors are more suitable to be used in surgical scene monitoring.

However, each 3D sensor has its own limitations. For example, Lidar (Light Detection and Ranging) can detect large-scale environment with high precision, but the point clouds or depth maps are very sparse. As for commodity RGBD sensors, such as Kinect, can accurately capture denser data, but limited in a small range from 0.5 to 4.5m. So, a proper method which can address these problems for fusing different modalities' data is important.

In this paper, we proposed a method which can fuse different modalities' 3D data to get a large-scale and dense point clouds. The key contributions of our work are as follows. First, we proposed a 3D data collecting system to reconstruct the medical scenes. By fusing the Lidar and Kinect data, a large-scale medical scene with more details can be reconstructed. Second, we proposed a location-based fast point clouds registration algorithm to deal with different modality datasets.

## 2. Related Works

3D point clouds reconstruction and registration are classic problems with a long history. After this process, a large point cloud with more information can be obtained. Registration can be divided into coarse registration and fine registration. Up to now, many coarse registration methods have been proposed, yet it still has much room to improve. In the Fast Point Feature Histogram (FPFH) [1], a histogram based descriptor is calculated for each point within the point cloud, over multiple scales. However, when the number of points is large, the calculation can become a serious problem. In the fine registration part, most widely used methods are ICP (Iterative Closest Point) [2] and its improved algorithms. These methods are able to optimize a coarse registration. However, it is hard to deal with the scenario where the initial distance between the two point clouds is too far. All of the above methods are designed for input point clouds pairs that are similar in order of magnitude and the same modality.

## 3. Methods

Our platform is built with Lidar and Kinect. The Lidar is used to detect large-scale scene, while the Kinect is used to capture small-scale but dense point clouds. An overview of our proposed registration method is shown in Fig. 1.

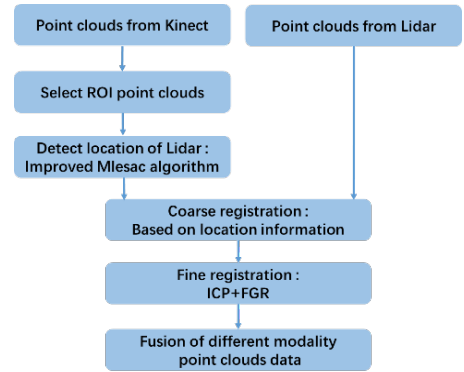

Fig. 1. Description of our point clouds registration method

### 3.1 Selection of ROI in Point Clouds

Because the points from Kinect are intensive, we need to reduce the points number to reduce the calculation. A ROI (Region of Interest) is selected by the value of axis Z. X, Y, Z are the coordinates of the points, Z is the height coordinate. And we choose the Z value between 0.8 to 1.8m as the ROI. Because this area contains the most important location information which we will use in the later registration part.

### 3.2 Point Clouds Coarse Registration

Registration task is usually divided into coarse registration and fine registration. As for the coarse part, calculating the relative location between Lidar and Kinect is a necessary step. Because the shape of Lidar is a cylinder, Mlesac (Maximum Likelihood Estimation Sample Consensus) [3] is used to detect the Lidar's location in the coordinate system of Kinect. In traditional Mlesac algorithm, the loss function is described as Eq. (1).

$$L_1 = \sum log\left(\gamma \left(\frac{1}{\sqrt{2\pi}\sigma}\right)^n \exp\left(-\left(\sum(\hat{x}_i^j - x_i^j)^2 + (\hat{y}_i^j - y_i^j)^2\right)/2\sigma^2\right) + (1-\gamma)\frac{1}{\nu}\right) \quad (1)$$

Where $\gamma \left(\frac{1}{\sqrt{2\pi}\sigma}\right)^n \exp\left(\left(\sum(\hat{x}_i^j - x_i^j)^2 + (\hat{y}_i^j - y_i^j)^2\right)/2\sigma^2\right)$ is the error model of Gaussian distribution, $(1-\gamma)\frac{1}{\nu}$ is the error model of uniform distribution, $\nu$ is the size of the searching window, and $\gamma$ is the mixture distribution factor of the two distributions.

Besides coordinates, Kinect can obtain 3D data with RGB information. So the RGB data should also be considered into the loss function. Equation (2) reflects the RGB loss.

$$L_2 \beta \sum ((R_{(x,y,z)i} - R_0)^2 + (G_{(x,y,z)i} - G_0)^2 + (B_{(x,y,z)i} - B_0)^2) \quad (2)$$

Where, $R_0, G_0, B_0$ are the colors of Lidar. Considering both coordinate and RGB information, the cost function can be improved and rewritten into Eq. (3). Then, by calculating the transform matrix from relative location, the coarse registration part is done.

$$L = L_1 + L_2 \quad (3)$$

### 3.3 Point Clouds Fine Registration

For the fine registration, we used ICP algorithm as our core method. To make the registration more accurate, FGR (Fast Global Registration) [4] is used combined with the ICP, which optimizes a robust objective defined densely over the surfaces. And due to this dense coverage, the algorithm directly produces an alignment that is as precise as which computed by well-initialized local refinement algorithms. As for ICP, suppose the source and target point clouds respectively are $P_t$ and $P_s$. $p_t^i$ is a certain point of set $P_t$, and $p_s^i$ belongs to $P_s$. The task can be defined as finding rotation matrix R and translation matrix T, in order to minimize the target function Eq. (4). For a certain $p_t^i$, we can find a point $p_s^i$, which is the closest point near $p_t^i$ in $P_s$. Temporary R and T can be easily calculated from $p_t^i$ and $p_s^i$. Using R and T, we can get a new pair of source and target point clouds. Then repeat the preceding procedure, until reaching the termination requirement.

$$f(R,T) = \frac{1}{N_p} \sum |p_t^i - R \cdot p_s^i - T|^2 \quad (4)$$

## 4. Experiments

### 4.1 Reconstruction of the Surgery Scene

In order to test our method in real environment, a surgery room is scanned as our object. As we can see in Fig. 3, the Lidar data is large-scale but sparse, the walls can be seen but the bed and person can hardly be recognized. From the Kinect data, more detail of the bed and person can be detected.

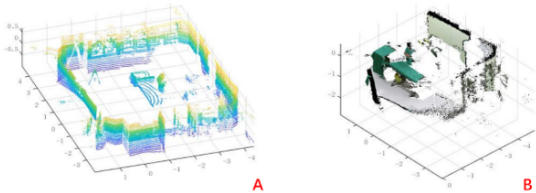

Fig. 3. Point Clouds from the surgery room. A is scanned by Lidar. B is scanned by Kinect.

### 4.2 Results of Registration

Using our proposed method, the surgery room is reconstructed with large-scale and abundant details. The final results are shown in Fig. 4. And Table 1 shows the accuracy of different registration algorithms. The distance between the corresponding feature points (manually select) are used as evaluation criterion. The first one is traditional ICP, where the second is ICP with location information, the location information can be seen as the coarse part. The third one is our proposed method. We add FGR to the fine registration part. The error has an obvious decline contrast to the former two.

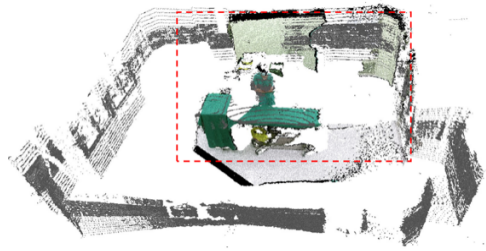

Fig. 4. The final results. The part in the red dashed line is detected by Kinect, and the rest is detected by Lidar.

Table 1 Accuracy of different registration algorithms.

| Algorithm | Error |
|---|---|
| ICP only | 7.43m |
| Location + ICP | 0.74m |
| Location+FGR+ICP | 0.11m |

## 5. Future Works

Although we have addressed the problems of large-scale medical scene reconstruction and point clouds registration from different modalities, there are still some limitations of our methods. The point clouds registration part is still time consuming which cannot satisfy the needs for clinical applications. In the future, we will try to use some deep learning methods on point clouds registration tasks, which could achieve a higher accuracy and less computation time.

## Acknowledgment

The authors acknowledge support from National Natural Science Foundation of China (81427803, 81771940), National Key Research and Development Program of China (2017YFC0108000), Soochow-Tsinghua Innovation Project (2016SZ0206), Beijing Municipal Natural Science Foundation (7172122, L172003).